\documentclass[11pt]{article}

\usepackage{acl}

\usepackage{times}
\usepackage{latexsym}

\usepackage[T1]{fontenc}

\usepackage[utf8]{inputenc}
\usepackage{amsmath}

\usepackage{tikz}
\usetikzlibrary{calc}
\usepackage{subcaption}
\usepackage{microtype}

\usepackage{inconsolata}

\usepackage{graphicx}

\usepackage{booktabs}
\title{C-World: A Computer Use Agent Environment Creator}
\usepackage{booktabs}   
\usepackage{makecell}   
\usepackage{graphicx}   
\usepackage{pifont}     
\usepackage{xspace}
\usepackage{xcolor}

\definecolor{geminiBG}{RGB}{210, 225, 245}   
\definecolor{gptBG}{RGB}{215, 235, 220}      
\definecolor{deepseekBG}{RGB}{245, 220, 190} 

\usepackage{tcolorbox}
\tcbuselibrary{skins}

\newcommand{\eg}{\emph{e.g.,}\@\xspace}

\newcommand{\cmark}{\textcolor{green!60!black}{\ding{51}}}
\newcommand{\xmark}{\textcolor{red!70!black}{\ding{55}}}

\newcommand{\rankup}[1]{\textcolor{teal}{\textbf{$\uparrow$#1}}}
\newcommand{\rankdown}[1]{\textcolor{purple}{\textbf{$\downarrow$#1}}}
\newcommand{\ranksame}{\textcolor{gray}{\textbf{--}}}

\author{
    Ziqiao Xi\thanks{\ \ Equal contribution} \quad
    Shuang Liang\footnotemark[1] \quad
    Qi Liu \quad
    Jiaqing Zhang \quad
    Letian Peng\\
    \textbf{Fang Nan} \quad
    \textbf{Meshal Nayim} \quad
    \textbf{Tianhui Zhang} \quad
    \textbf{Rishika Mundada}\\
    \textbf{Lianhui Qin} \quad
    \textbf{Biwei Huang} \quad
    \textbf{Kun Zhou}\thanks{\ \ Corresponding author: \href{mailto:franciskunzhou@gmail.com}{franciskunzhou@gmail.com}}\\
        UC San Diego \\
}

\begin{document}
\maketitle

\begin{abstract}
To close the gap between LLM-based agents and humans in planning and reasoning, agents need large-scale, diverse environments for continuous learning---yet building such environments is itself prohibitively expensive. We present C-World, an environment creation system that enables users to build agent environments on demand. We define a complete agent environment through four components: an Action Space of 5,571 format-unified tools across 204 common applications, a Task Distribution engine that synthesizes long-horizon workflows with wild constraints, a Transition Function implemented as a state controller that injects realistic failures and perturbations, and a Reward Signal combining verifiable metrics with LLM-based judgment. C-World operates in two modes: a realistic mode grounded in live API execution, and a synthesized mode powered by the World Engine, which approximates tool behavior without live service access, enabling scalable environment creation---including environments for domains and tools that do not yet exist in the real world. Evaluation of nine state-of-the-art LLMs reveals that planning ability is uniformly strong but execution remains the bottleneck, and that constraint following---not tool invocation---is the dominant failure mode. The World Engine achieves Spearman $\rho = 0.883$ ranking correlation with real execution, and fine-tuning on just 1,170 C-World trajectories outperforms baselines trained on 119k samples, demonstrating C-World's dual value as a rigorous evaluation environment and a scalable data engine. Our code and data are available at \url{https://ziqiao-git.github.io/C-World/}.
\end{abstract}

\section{Introduction}

Owing to scaling laws, large language models (LLMs) have exhibited strong reasoning and planning abilities \citep{kaplan2020scaling, wei2022chain, bubeck2023sparks}, catalyzing the rapid rise of autonomous LLM-based agents \citep{xi2023rise, park2023generative}. In open-world settings, such agents are expected to decompose complex goals into multi-step plans, coordinate diverse sub-tasks, and adapt their strategies as situations evolve. Tool use—querying services, operating software, and executing workflows—serves as the primary interface through which agents act on the world. In its narrow sense, tool use is approaching a solved problem: frontier LLMs already achieve near-saturated performance on single-turn function calling and schema compliance benchmarks \citep{patilberkeley}. 

Despite rapid progress, a significant gap remains between LLM-based agents and humans in planning and reasoning for real-world tasks. When faced with complex objectives involving large decision spaces, underspecified requirements, and many interacting constraints, current agents frequently fail to maintain coherent long-horizon plans, adapt to evolving situations, or recover from intermediate errors \citep{zhouwebarena, mialon2023gaia, liu2023agentbench}. Humans, by contrast, develop these abilities naturally—through daily life tasks, education, games, and countless other experiences, both manually crafted and naturally occurring. This continuous exposure to diverse, open-ended challenges is what builds the robust planning and adaptive reasoning that agents currently lack. The implication is clear: to close this gap, agents need not just better models, but a large-scale environment that can supply the same breadth and density of learning experiences.

However, building such an environment is itself a major challenge. Existing agent environments are neither sufficiently diverse nor easy to create: most are hand-crafted for narrow domains, and expanding them to new tools, tasks, or failure modes requires substantial human effort and engineering cost \citep{liu2023agentbench, zhouwebarena}. As shown in Table~\ref{tab:apps_tools_rounds}, current benchmarks are bottlenecked by limited tool coverage, simplified settings, and static task sets that cannot evolve. Scaling up environment creation through brute-force manual curation is prohibitively expensive, yet diversity and scale are precisely what agents need to develop robust planning and reasoning. To overcome this, we take a fundamental shift: rather than building a fixed benchmark, we aim to build an environment creation system—one that, given a set of seed tasks, tools, and evaluation criteria, can automatically grow and evolve into increasingly diverse and complex environments. Our goal is to enable users to create any environment they need, whether for specialized domain testing or personalized training, without starting from scratch each time.

\begin{figure}[t]
    \centering
    \includegraphics[width=0.9\linewidth]{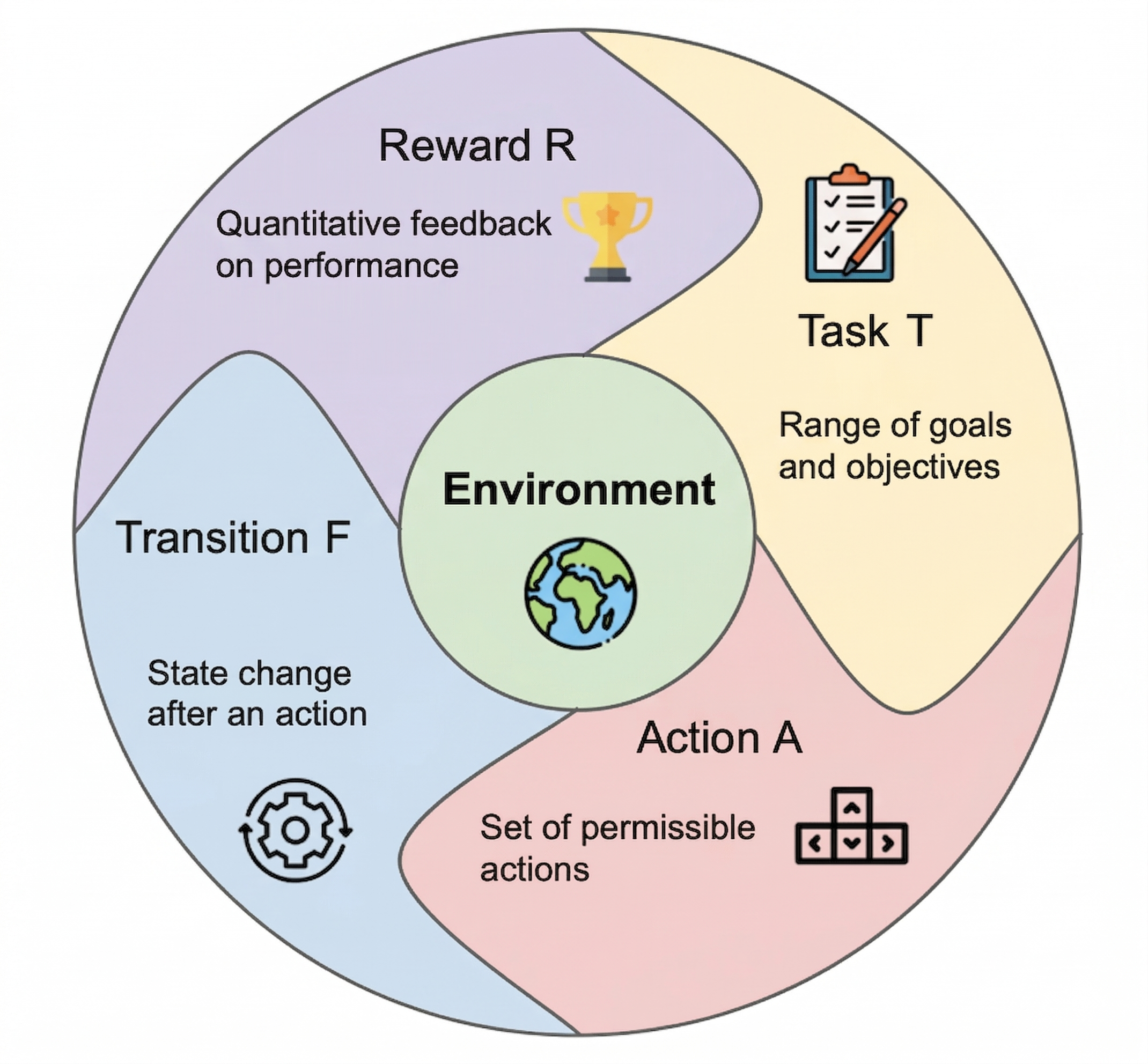}
    \caption{We define an agent environment through four essential components: Action Space ($\mathcal{A}$), Task Distribution ($\mathcal{T}$), Transition Function ($\mathcal{F}$), and Reward Signal ($\mathcal{R}$).}
    \label{fig:env-components}
\end{figure}

We argue that four essential components are required (Figure~\ref{fig:env-components}): (i) an \emph{Action Space}~($\mathcal{A}$)---a large-scale, format-unified library of executable tools that reflects the breadth of real-world services; (ii) a \emph{Task Distribution}~($\mathcal{T}$)---a scalable mechanism for generating diverse, long-horizon tasks with complex, compositional constraints; (iii) a \emph{Transition Function}~($\mathcal{F}$)---realistic state transitions upon tool execution, including not only normal responses but also failures, timeouts, and state drift that agents must handle; and (iv) a \emph{Reward Signal}~($\mathcal{R}$)---an automated, multi-dimensional evaluation protocol that provides reliable feedback on agent behavior. As shown in Table~\ref{tab:apps_tools_rounds}, most existing efforts are designed as static benchmarks for one-time evaluation, rather than interactive environments for continuous agent learning, and none incorporates a world model that enables scalable training without live service dependencies.

\begin{table*}[t]
\centering
\small
\setlength{\tabcolsep}{4pt}
\begin{tabular}{lrrrcccc}
\toprule
& \makecell{\# Apps / \\  \# Servers} &
\makecell{\# Real / \\ Tools} &
\makecell{\# Avg. / \\ Rounds} &
\makecell{Env.-level \\ Synthesis} &
\makecell{Wild \\ Constraints} &
\makecell{State\\ Controller} &
\makecell{Fuzzy \\ Instructions} \\
\midrule
BFCL v4~\citep{patilberkeley}      & 1 / None & 2000+ & 3.8 & \xmark & \xmark & \xmark & \xmark \\
ToolBench~\citep{qin2023toolllm} & N/A & 3451 & 3.7 & \cmark & \xmark & \xmark & \xmark \\
AgentBench~\citep{liu2023agentbench} & N/A    & 18 & 11.9 & \xmark & \xmark & \xmark & \xmark \\
ToolEyes~\citep{ye2024tooleyes}  & N/A & 568  & 4.5 & \xmark & \xmark & \xmark & \xmark \\
StableToolBench~\citep{guo2024stabletoolbench} & N/A & 3451 & 3.7 & \cmark & \xmark & \cmark & \xmark \\
MCPEval~\citep{liu2025mcpeval}          & 8 / 14 & 121 & N/A & \cmark & \xmark & \xmark & \xmark \\
LiveMCPBench~\citep{mo2025livemcpbench} & 36 / 70 & 527 & 5.6 & \xmark & \xmark & \xmark & \xmark \\
MCP-Universe~\citep{luo2025mcp} & 8 / 11 & 133 & 7.5 & \xmark & \xmark & \xmark & \xmark \\
MCP-Bench~\citep{wang2025mcp} & 20 / 28  & 250  & 44.2  & \cmark & \cmark & \xmark & \cmark \\
Toolathlon~\citep{li2025tooldecathlonbenchmarkinglanguage} & 17 / 32  & 604  & 26.8  & \xmark & \cmark & \cmark & \cmark \\
ToolAce~\citep{liu2024toolace}   & N/A & 26507 & 1.6 & \cmark & \xmark & \xmark & \xmark \\
Toucan~\citep{xu2025toucan}      & 179 / 495 & 2000+ & 2 & \cmark    & \xmark    & \xmark  & \xmark \\
\midrule
\textbf{C-World (Ours)} & 204 / 276 & 5,571 & 28.5 & \cmark & \cmark & \cmark & \cmark \\
\bottomrule
\end{tabular}
\caption{Comparison of related work along environment design axes: \# Apps, \# Tools, and \# Rounds denote the average number of covered real-world apps, available tools and interaction rounds per task. We additionally mark whether the system supports environment-level synthesis for scaling up, wild constraints, flexible state controlling, and fuzzy instructions.}
\label{tab:apps_tools_rounds}
\end{table*}

To this end, we build C-World in two layers: a realistic foundation grounded in 5,571 live MCP tools with a state controller for failure injection, and a World Engine that synthesizes new environments without live service access---even for tools and domains that do not yet exist in the real world. Using C-World, we evaluate state-of-the-art LLMs and train smaller models, yielding the following findings:

\noindent\textbf{Evaluation findings:}
\begin{itemize}
\item All LLMs exhibit strong planning ability, but execution ability lags significantly behind, causing the dominant gap in task success rate;
\item Constraint following, rather than tool invocation, is the primary failure mode across models;
\item DeepSeek-v3.2 achieves the best robustness, with stronger recovery from intermediate unpredictable disruptions;
\item Higher tool-call volume does not imply higher success—weaker models make more calls yet fail more due to poor reasoning.
\end{itemize}

\noindent\textbf{Training findings:}
\begin{itemize}
\item Fine-tuning Qwen2.5-7B and Qwen3-8B on just 1,170 trajectories collected from C-World outperforms baselines trained on 119k samples, demonstrating the environment's data efficiency;
\item The World Engine achieves $\rho$ = 0.883 ranking correlation with real execution, confirming that simulated environments can reliably substitute live APIs for scalable training.
\end{itemize}




\section{C-World}
\label{sec:C-World}
\begin{figure*}[t]
    \centering
    \includegraphics[width=\linewidth]{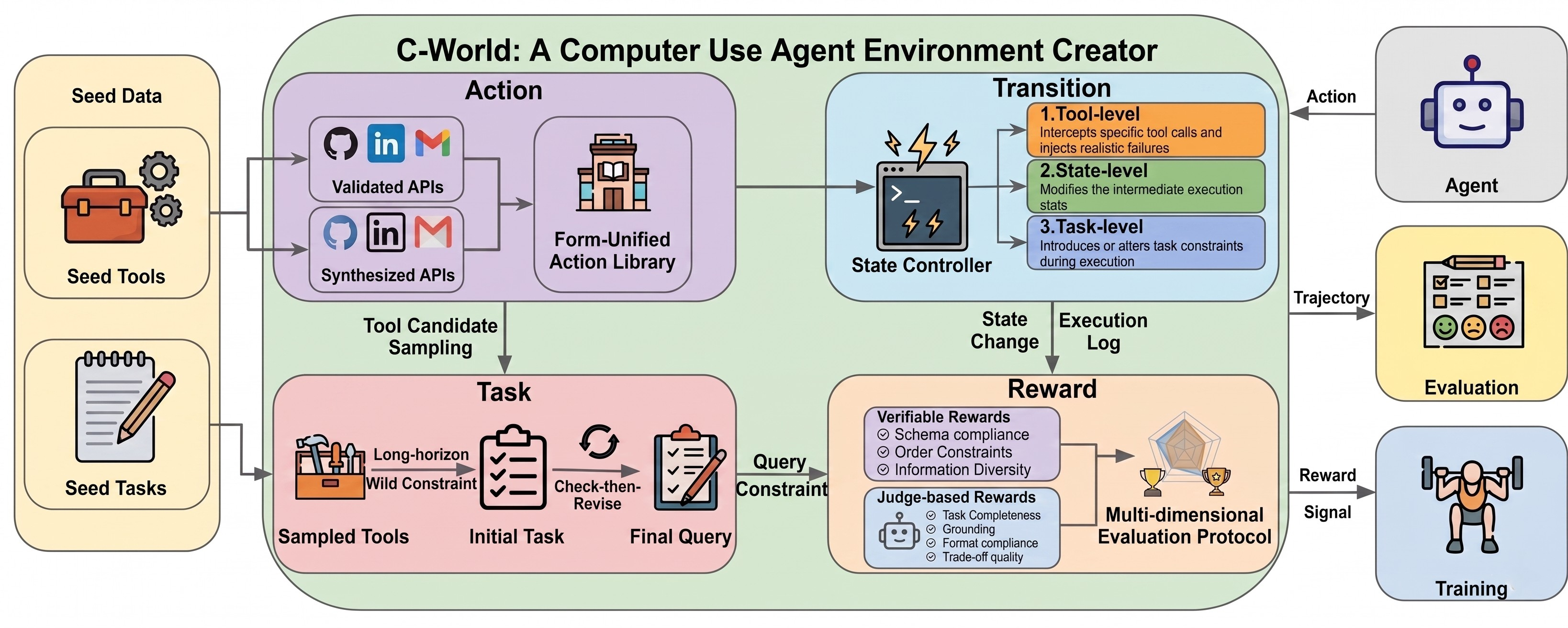}
    \caption{The \textsc{C-World} framework, defining an agent environment through four components: \textbf{Action} (\S\ref{sec:action-space})---a unified library of validated and synthesized APIs; \textbf{Task} (\S\ref{sec:task-distribution})---long-horizon queries with wild constraints synthesized via a check-then-revise loop; \textbf{Transition} (\S\ref{sec:transition})---a State Controller injecting tool-, state-, and task-level disruptions; and \textbf{Reward} (\S\ref{sec:reward})---a multi-dimensional protocol combining verifiable and judge-based signals. The Agent interacts via the State Controller, producing trajectories for both evaluation and training.}
    \label{fig:pipeline}
\end{figure*}

We define an agent environment as comprising four essential components:
(i)~an \emph{Action Space}~$\mathcal{A}$---a large-scale, format-unified library of executable tools that an agent can invoke;
(ii)~a \emph{Task Distribution}~$\mathcal{T}$---a scalable mechanism for generating diverse tasks with varying complexity, constraints, and long-horizon dependencies;
(iii)~a \emph{Transition Function}~$\mathcal{F}$---the dynamics that determine how the environment state evolves after each action, encompassing both normal responses and realistic disruptions such as failures, timeouts, and state drift;
and (iv)~a \emph{Reward Signal}~$\mathcal{R}$---an automated, multi-dimensional evaluation protocol that provides reliable feedback on agent behavior.
A complete environment must integrate all four components to support both rigorous evaluation and scalable training.
Additionally, C-World provides a World Engine that approximates~$\mathcal{F}$ without live API access, enabling large-scale environment creation beyond what real-world services alone can support~(\S\ref{sec:world-engine}).

\subsection{Action Space: Extensible Toolset}
\label{sec:action-space}
 
The action space~$\mathcal{A}$ defines what an agent can do. We build it to be large-scale, realistic, and extensible by unifying tools under the Model Context Protocol~(MCP), selecting commonly used tools in daily human workflows, and validating them under controlled credentials. To support open-world scenarios where no tool list is given in advance, we additionally provide a retrieval index that enables agents to discover and load tools on demand from the full library.

\paragraph{MCP Server Selection.}
We construct an executable tool universe by curating 5,571 tools from 276 MCP servers on Smithery\footnote{\url{https://smithery.ai}}, covering 204 production-grade applications (\eg Gmail, GitHub). Our selection follows a simple principle: we prioritize tools that are \emph{commonly used}, \emph{general-purpose}, and \emph{popular} in everyday human work, so that the resulting toolset reflects realistic workflows (\eg documents, coding, search, and collaboration). We build this collection through a hybrid pipeline that combines automated registry crawling with manual augmentation, allowing us to scale coverage while preserving tool diversity and practical relevance.
To keep the setting faithful to real-world APIs, we retain each tool’s native invocation interface and directly use its original argument schema. For servers requiring authentication, we provision dedicated virtual accounts without personal information and configure them with controlled credentials, enabling authorized execution while ensuring reproducibility across runs.

\paragraph{Tool Validation.}
The availability of the registry does not guarantee the executability of the tools. After filtering servers that fail basic MCP handshakes, we validate each individual tool under the authenticated evaluation personas using a three-stage pipeline: (i) \emph{authenticated availability} ensuring the server can list and access the tool under the persona; (ii) \emph{successful invocation} verifying the tool runs with schema-respecting arguments; and (iii) \emph{usable responses} filtering tools that execute but return error payloads or non-actionable outputs. This process yields a reliable \emph{working} toolset that can be safely used for downstream task synthesis and trajectory-based evaluation.

\paragraph{Tool Retrieval Index.}
Given the scale of our open-world tool registry, agents need an efficient mechanism for tool discovery, analogous to a search engine over APIs. We therefore index all validated tools as vector documents by concatenating their server identity, tool name, natural-language description, and schema signature. We embed these documents using \textsc{BGE-M3}~\citep{chen-etal-2024-m3} and store them in a FAISS index~\citep{8733051}. At runtime, we provide a lightweight MCP service \texttt{search\_tools(query, k)} on a locally deployed server, enabling agents to issue natural language queries and retrieve the top-$k$ relevant tools for on-demand loading and invocation.

\subsection{Task Distribution: Task Creation Engine}
\label{sec:task-distribution}

The task distribution~$\mathcal{T}$ determines what challenges agents face. Our goal is to generate tasks that resemble what humans encounter in daily work: they are \emph{long-horizon}, require \emph{coordinated multi-step execution}, and include \emph{wild constraints} (\eg multi-source verification and explicit trade-offs). To this end, we first sample a coherent and diverse candidate set of tools that can plausibly serve one workflow, and then synthesize a task that naturally uses these tools while incorporating diverse constraints.

\paragraph{Tool Candidate Sampling.}
We build a tool candidate set that is both semantically coherent (so the task is natural) and diverse across services (so it requires real orchestration). Concretely, we first randomly sample 1--3 \emph{seed tools}, then form a retrieval query from their names and descriptions, and use our tool retrieval index to recall a larger set of semantically related tools. Next, we group recalled tools by their underlying servers/apps and uniformly sample across these groups (round-robin) to assemble the final candidate set. This step prevents the candidate set from being dominated by a single service and encourages cross-app workflows. We also control difficulty by sampling the final bundle size from a configurable range and enforcing a minimum number of distinct servers.

\paragraph{Synthesizing Tasks with Constraints.}
Given the selected tool candidate set, we first synthesize an initial task query by prompting an LLM to produce a task instruction with a unified goal that plausibly requires composing the candidate tools. Starting from this draft, we run an iterative \emph{check-then-revise} loop to progressively improve realism and increase constraint density. In each round, we automatically evaluate the draft from two perspectives:
(i)~\emph{tool coverage}: whether the request legitimately motivates using all or most candidate tools (supported by per-tool rationales);
(ii)~\emph{constraint quality}: whether the constraints are diverse, non-trivial, and interact in ways that create long-horizon dependencies.
If either criterion is not met, we generate targeted feedback and revise the prompt to re-synthesize a stronger query in the next iteration. The final task instruction includes the user query, a set of constraints, and grounded rationales describing each tool's role, enabling long-horizon supervision and reliable evaluation.

\paragraph{Preventing Retrieval Shortcuts.}
To prevent tasks from being trivially solvable by keyword matching against our synthesis queries, we: (i) apply \emph{fuzzy rewriting} that abstracts tool names into intent-level descriptions (\eg ``send the summary to the team'' instead of ``use \texttt{slack\_post\_message}''); (ii) expose only the \texttt{search\_tools} interface at evaluation time, forcing tool discovery against the full 5,571-tool index; and (iii) require task decomposition and targeted sub-queries, since long-horizon tasks cannot be solved by a single query. These ensure task difficulty reflects genuine planning rather than synthesis artifacts.
\subsection{Transition Function: State Controller}
\label{sec:transition}

The transition function~$\mathcal{F}$ governs how the environment state evolves after each agent action. In real-world settings, this includes not only successful responses but also the wide variety of failures and perturbations that agents must handle. To faithfully model these dynamics, we implement~$\mathcal{F}$ through a \emph{state controller}: a middleware layer that sits between the agent and the tool execution backend. The state controller serves two purposes: it forwards actions to real (or simulated) services to produce normal responses, and it can inject controlled disruptions according to pre-defined policies to stress-test agent robustness. Concretely, the state controller is implemented as a lightweight Python middleware inside the agent runtime, interposed between the Actor and the MCP client. It intercepts outgoing tool calls and incoming observations, applying intervention policies before the observation is returned to the agent. This in-runtime design avoids the overhead of an HTTP proxy and keeps intervention logic fully reproducible across runs. Concrete examples of each intervention type are provided in Appendix~\ref{app:state-controller-examples}.

Unlike random noise injection, the state controller applies \emph{targeted} interventions that are reproducible and specifically designed to probe recovery and adaptation. We categorize interventions into three complementary types:

\begin{itemize}
    \item \textbf{Tool-level control:} Intercepts specific tool calls and injects realistic failures, \eg timeouts, rate limits, and temporary unavailability.
    \item \textbf{State-level control:} Modifies the intermediate execution state exposed to the agent, \eg delayed or corrupted results, changed service state (\eg session timeout), emulating common real-world drift and instability.
    \item \textbf{Constraint-level control:} Introduces or alters task constraints during execution (\eg a new preference or a changed deadline), testing whether the agent can re-plan and re-optimize under evolving requirements.
\end{itemize}

\subsection{Reward Signal: Evaluation Protocol}
\label{sec:reward}

The reward signal~$\mathcal{R}$ provides the feedback that makes an environment useful for both evaluation and training. Our evaluation protocol combines two complementary types of reward signals to cover the full spectrum of agent behavior.

\paragraph{Verifiable Rewards.}
For properties that admit deterministic checking, we compute rewards directly from execution logs without model-based judgment. These include \emph{schema compliance} (validating every tool call against official JSON schema definitions), \emph{order constraints} (comparing execution timestamps against the task's dependency graph), and \emph{information diversity} (counting distinct servers or sources accessed). Such signals are cheap, reproducible, and free from judge noise.

\paragraph{Judge-based Rewards.}
For semantic properties that require understanding of intent and context, we employ an LLM-as-judge protocol with three frontier models (GPT-4o, GPT-5.1, DeepSeek-V3.2) aggregated via majority vote. Judges score \emph{task completeness}, \emph{grounding}, \emph{format compliance}, and \emph{tradeoff quality} from the agent's trajectory and final answer; full definitions are in Appendix~\ref{app:eval-metrics}.

\section{Using C-World}
\label{sec:using-cworld}

C-World supports two complementary modes of environment creation, each serving different needs in the agent development lifecycle.

\subsection{Realistic Environments}
\label{sec:realistic-env}

In the realistic mode, agents interact with real-world services through live API calls, producing fully faithful execution trajectories. The task creation engine (\S\ref{sec:task-distribution}) synthesizes complex, multi-step workflows grounded in real tools, and the state controller (\S\ref{sec:transition}) can inject controlled disruptions---such as timeouts, state drift, and evolving constraints---to stress-test robustness beyond the idealized happy path. The evaluation protocol (\S\ref{sec:reward}) provides both verifiable and judge-based reward signals on the resulting trajectories.

This mode is best suited for high-fidelity evaluation, where faithfulness to real service behavior is essential. It produces ground-truth trajectories that serve as reliable benchmarks for comparing agent capabilities, and also provides the real demonstrations needed to bootstrap the World Engine (\S\ref{sec:world-engine}).

\subsection{Synthesized Environments}
\label{sec:synthesized-env}

While realistic environments provide faithful evaluation, they are bottlenecked by live API costs, rate limits, and service instability. To enable environment creation at scale, C-World provides a second mode powered by the \emph{World Engine}, which approximates the transition function~$\mathcal{F}$ without live API access (see Figure~\ref{fig:pipeline}).

\paragraph{World Engine.}
\label{sec:world-engine}
Given a tool call~$a_t$ and execution context~$s_t$, the World Engine predicts a realistic tool response~$o_{t+1} = f(s_t, a_t)$. We organize tools into functional categories (\eg email, calendar, code hosting) and manually construct a \emph{category-level card} for each group, encoding shared response patterns, field structures, and common failure modes, along with few-shot demonstrations. Conditioned on the card, tool schema, and a session-level execution log for state consistency, the World Engine generates realistic responses and generalizes to unseen tools within the same category without per-tool demonstrations. Because it operates on schema and category cards alone, the World Engine can also synthesize entirely new environments---enterprise agents, targeted stress-test scenarios (\eg periodic timeouts or token expiration), or domain-specific training grounds (\eg medical or legal databases)---without live service dependencies.

\section{Agent Framework}
\label{sec:agent-framework}

Long-horizon tasks in C-World require agents to coordinate many interdependent sub-goals across dozens of steps. Purely zero-shot execution often fails due to premature termination and poor error recovery. As shown in Table~\ref{tab:turn1_statistics}, even models that perform well in the first turn can degrade significantly over long horizons without sustained guidance, motivating the need for explicit planning and feedback throughout execution. To this end, we adopt a \emph{planner--actor} decomposition where both roles are instantiated by the same LLM.

The \textbf{Planner} receives the task instruction and decomposes it into a reference sub-goal graph before execution begins. During execution, it tracks progress by comparing the Actor's trajectory against this graph, intervening with corrective feedback when the Actor deviates or stalls.

The \textbf{Actor} operates step-wise following the ReAct paradigm~\citep{yao2022react}: it reasons about the current state, queries the tool retrieval index to select relevant tools, invokes them with appropriate arguments, and observes the result before proceeding to the next action.

After each execution step, the Planner verifies the result against the current sub-goal and marks it as \emph{Complete} or \emph{Pending}. The trajectory terminates when all sub-goals are complete, or when a maximum turn limit or repeated inaction limit is reached.

\section{Experiments}

\label{sec:experiments}
We conduct experiments to validate C-World along three axes: (1)~as a realistic evaluation environment that reveals meaningful capability differences across models, (2)~as a scalable environment via the World Engine that faithfully preserves model rankings without live APIs, and (3)~as a data engine whose trajectories enable efficient agent training.

\begin{table*}[t]
\centering
\small
\setlength{\tabcolsep}{4pt} 
\begin{tabular}{lcccccccc}
\toprule
& & \multicolumn{2}{c}{Quality} & \multicolumn{2}{c}{Robustness} & \multicolumn{1}{c}{Constraint} & \multicolumn{2}{c}{Planning} \\
\cmidrule(lr){3-4} \cmidrule(lr){5-6} \cmidrule(lr){7-7} \cmidrule(lr){8-9}
Model & Overall & Comp. & Grnd. & Recov. \% & Flex. \% & Format \% & \# Calls & Decomp. \\
\midrule
gemini-3-pro-preview & \textbf{5.87} & \textbf{4.75} & 2.58 & 89.0 & 68.8 & 53.9 & 47.9 & \textbf{8.66} \\
claude-opus-4.5      & 5.42 & 4.70 & 2.93 & 83.7 & 60.8 & 51.0 & 45.2 & 7.72 \\
deepseek-v3.2        & 4.97 & 4.00 & 2.18 & \textbf{90.6} & \textbf{72.4} & 39.5 & 21.7 & 8.04 \\
glm-4.6v             & 4.86 & 4.01 & 1.18 & 71.5 & 57.3 & 34.2 & 18.0 & 8.50 \\
grok-4               & 4.78 & 3.80 & 1.95 & 89.0 & 63.6 & \textbf{68.3} & 27.4 & 8.28 \\
gpt-oss-120b         & 4.66 & 3.42 & 1.28 & 72.7 & 59.7 & 35.8 & 14.4 & 8.10 \\
gpt-5.2              & 4.43 & 3.42 & \textbf{3.80} & 79.3 & 55.4 & 12.4 & 29.2 & 7.73 \\
qwen3-235b-a22b      & 3.53 & 2.56 & 1.17 & 88.1 & 66.1 & 31.3 & 11.2 & 8.51 \\
gpt-4o-mini          & 3.07 & 1.13 & 0.85 & 50.6 & 39.7 & 3.3  & 51.7 & 7.71 \\
\bottomrule
\end{tabular}
\caption{Main Leaderboard Summary. We report the Overall Score alongside key metrics: \textbf{Quality} (Completeness \& Grounding), \textbf{Robustness} (Recovery Rate \& Flexibility), \textbf{Constraint} (Format), and \textbf{Planning} (Avg. Tool Calls \& Goal Decomposition). See Appendix~\ref{app:detailed_results} for the full breakdown.}
\label{tab:leaderboard_summary}
\end{table*}

\subsection{Results Analysis}
Table~\ref{tab:leaderboard_summary} presents the comparative performance across different model tiers. The details of evaluation metrics are in Appendix.
Based on the metrics, we observe systematic performance differences across models, accompanied by distinct behavioral patterns.

\paragraph{Overall Performance}

Models exhibit clear performance differences across tiers. Frontier models such as gemini-3-pro-preview and claude-opus-4.5 consistently dominate, achieving top-tier Answer Quality ($>4.70$ Completeness) and Success Rate ($>88\%$). Notably, deepseek-v3.2 emerges as a strong open-weight contender, attaining the highest Recovery Rate (90.6\%) and Flexibility (72.4\%), even surpassing proprietary models in robustness. In contrast, gpt-4o-mini struggles in open-world settings, posting the lowest Completeness (1.13) and Grounding (0.85), indicating limited readiness for complex autonomous workflows.

In comparison, \texttt{gpt-5.2} shows signs of reduced stability when coordinating long-horizon tasks.
Despite possessing the strongest Grounding (3.80) among other models, it falters significantly in task completion.
Qualitative analysis attributes this decline to a pattern of early abandonment: as early as the second turn, the model retreats into passivity, discarding systematic planning and rigorous search.
Instead of actively navigating the tool space, it resorts to hallucinating non-existent interfaces to bypass complex sub-tasks.
Consequently, critical requirements remain unsatisfied, resulting in a severely penalized Completeness score.

\paragraph{Behavioral Analysis}

Beyond simple rankings, analyzing the sub-metrics reveals critical insights into how models fail or succeed.

Higher tool-using rate does not imply higher success rate: gpt-4o-mini exhibits the highest volume of Tool Calls (51.71) and \# Turns (6.45), yet yields the lowest output quality. This inverse correlation indicates a failure pattern: the agent falls into a ``looping'' trap, repeatedly invoking tools without effectively synthesizing observations, whereas gemini-3-pro-preview converts similar high activity (47.86 calls) into superior completeness (4.75).

All LLMs exhibit strong planning ability, but their execution ability is not aligned and causes a significant gap in task success rate. Across the board, Goal Decomposition scores are consistently high (7.7--8.6), proving that most models function as competent Planners. However, the divergence in Tool Calls and Success Rate shows that the bottleneck lies in the Actor's endurance. For instance, qwen3-235b-a22b plans well (8.51) but executes poorly (11.15 calls, 2.56 completeness), failing to sustain the necessary action sequence.

\paragraph{Planning vs. Endurance.}
Table~\ref{tab:turn1_statistics} contrasts initial action against final standing to decouple reasoning from resilience. Because the Planner's intervention only begins after Turn~1, the Turn-1 row corresponds to a standard monolithic ReAct baseline: the Actor operates on the raw user query without plan decomposition or progress feedback. The final Completeness column then reflects sustained execution under our full Planner--Actor framework, making this table a direct ablation of the two-role decomposition.
Despite the strong first-turn score, gpt-5.2 collapses the most over time, dropping five ranks, showing that doing well at the start does not mean the model can sustain performance over longer interactions.
Conversely, gemini-3-pro-preview and glm-4.6v overcome mediocre starts to climb +4 and +5 positions respectively, proving that in open-world settings, long-term reflection is more critical than a strong start.
Deepseek-v3.2 strikes the best balance, maintaining high quality with zero rank drift and minimal steps.

\begin{table}[t]
\centering
\footnotesize
\setlength{\tabcolsep}{4pt}
\begin{tabular}{lccc}
\toprule
Model & \makecell{Avg. Steps\\(Turn 1)} & \makecell{Turn 1\\ Result} & \makecell{Rank Shift\\(long horizon)} \\
\midrule
gpt-5.2              & 20.22 & \textbf{5.00} & \rankdown{5} \\
claude-opus-4.5      & 30.20 & 4.96 & \ranksame \\
grok-4               & 25.79 & 4.72 & \rankdown{2} \\
deepseek-v3.2        & \textbf{10.78} & 4.67 & \ranksame \\
gemini-3-pro         & 33.19 & 4.62 & \rankup{4} \\
qwen3-235b-a22b          & 10.75 & 4.56 & \rankdown{2} \\
gpt-oss-120b         & 12.61 & 4.03 & \ranksame \\
glm-4.6v             & 16.59 & 3.55 & \rankup{5} \\
gpt-4o-mini          & 22.50 & 3.04 & \ranksame \\
\bottomrule
\end{tabular}
\caption{First-turn dynamics vs. long-horizon outcome, serving as an ablation of the Planner--Actor decomposition. \textbf{Turn 1 Result} corresponds to a monolithic ReAct baseline (no Planner intervention yet), while \textbf{Rank Shift} tracks how each model's standing changes once the full Planner--Actor loop takes effect. \textcolor{teal}{$\uparrow$} indicates models that gain over long horizons under Planner guidance; \textcolor{purple}{$\downarrow$} indicates models that lose ground due to instability.}
\label{tab:turn1_statistics}
\end{table}
\subsection{Human Alignment Measurement}
Our LLM-as-judge protocol achieves human-level reliability on $N=40$ queries: DeepSeek-V3.2 attains $\rho=0.759$ and GPT-5.1 attains $\rho=0.733$ Spearman correlation with human rankings, closely approaching the human--human ceiling of $\rho=0.773$. The cross-family majority-vote ensemble also neutralizes single-developer bias: GPT-4o-mini ranks last overall and GPT-5.2 is outperformed by non-OpenAI models despite two OpenAI judges in the ensemble. Full alignment analysis is deferred to Appendix~\ref{app:human-alignment}.
\subsection{Characteristic Analysis}
Beyond aggregate metrics, we map execution logs to five anthropomorphic dimensions (Diligence, Prudence, Grit, Introspection, Strategic) to characterize model personas. Figure~\ref{fig:radar-3models} shows distinct styles across representative models; full dimension definitions are in Appendix~\ref{app:personas}.

\newcommand{\modelpersonatag}[3]{%
  \tcbox[
    colback=#3,
    colframe=#3,
    arc=3pt,
    boxsep=1pt,
    left=4pt,
    right=4pt,
    top=1pt,
    bottom=1pt
  ]{%
    \begin{tabular}{@{}c@{}}
      {\scriptsize\bfseries #1} \\[-2pt]
      {\tiny\itshape #2}
    \end{tabular}
  }%
}

\newcommand{\persona}[3]{%
  \parbox[t]{0.95\linewidth}{%
    \centering
    \modelpersonatag{#1}{#3}{#2}%
  }%
}

\begin{figure}[t]
  \centering
  \setlength{\tabcolsep}{3pt}
  \begin{tabular}{@{}c c c@{}}

    \begin{minipage}[t]{0.33\linewidth}
      \centering
      \includegraphics[width=\linewidth]{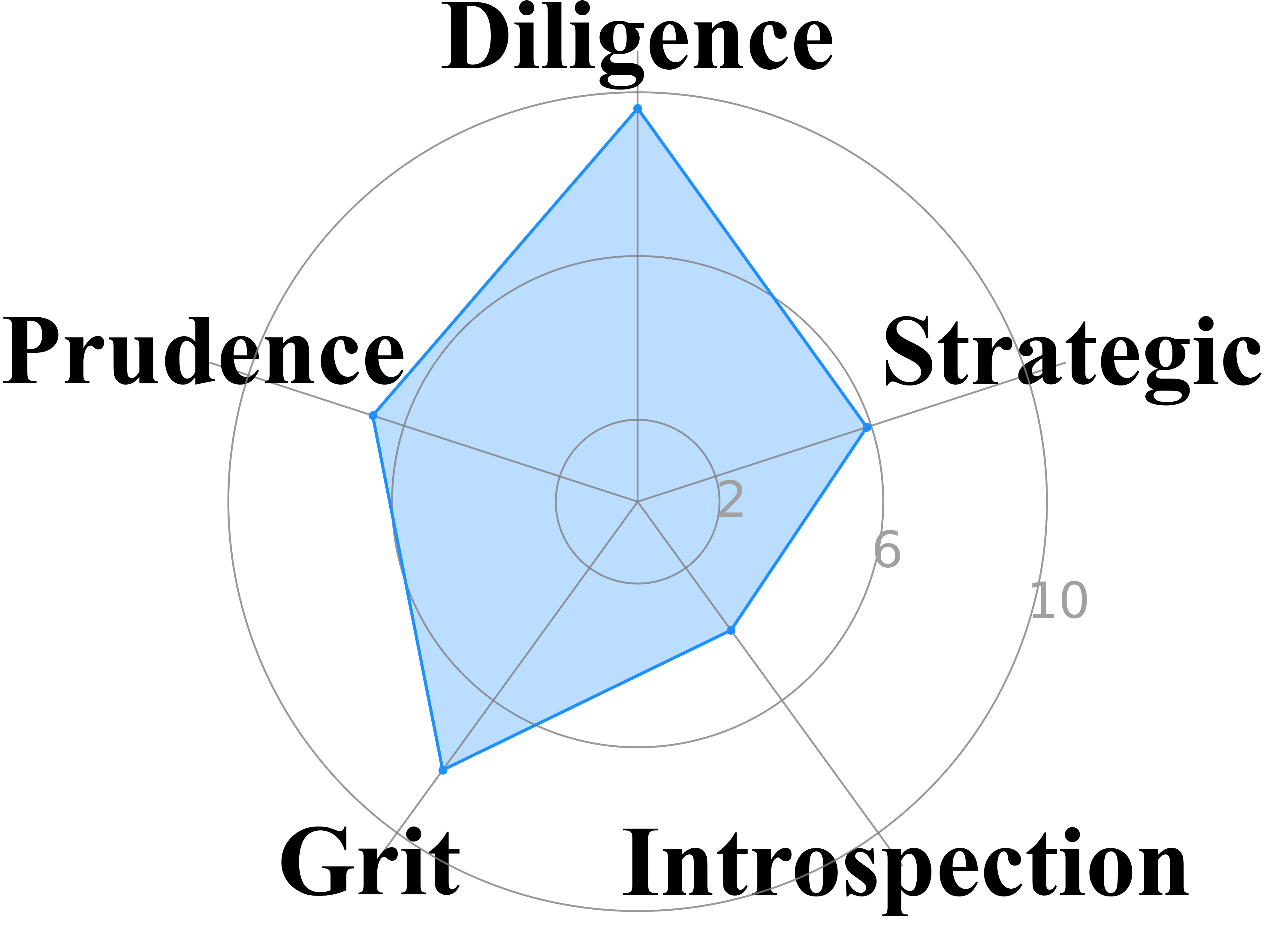}\vspace{1pt}
      \persona
        {Gemini-3-Pro}
        {geminiBG}
        {meticulous scholar}
    \end{minipage}
    &
    \begin{minipage}[t]{0.33\linewidth}
      \centering
      \includegraphics[width=\linewidth]{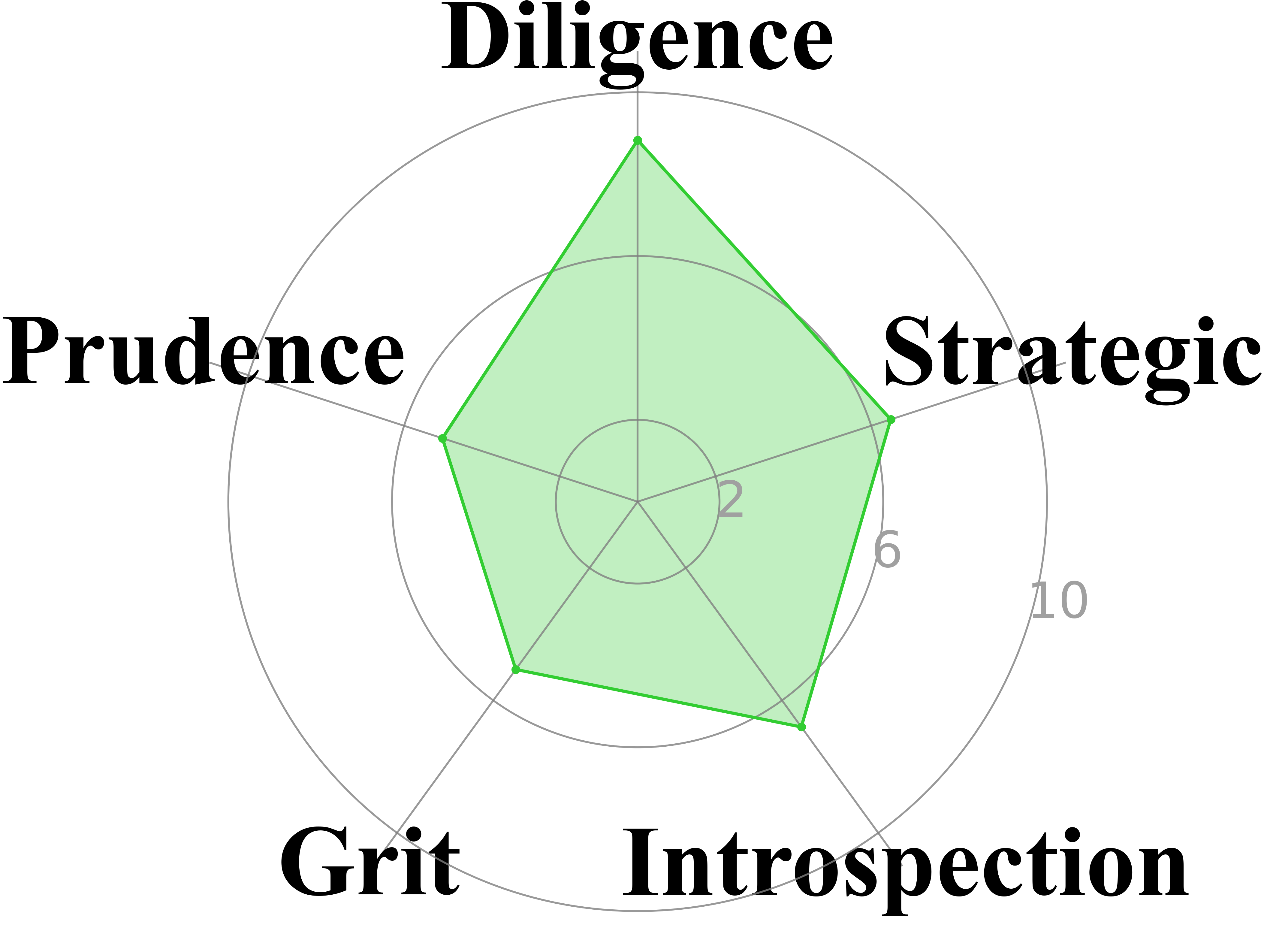}\vspace{1pt}
      \persona
        {GPT-4o-mini}
        {gptBG}
        {bold worker}
    \end{minipage}
    &
    \begin{minipage}[t]{0.33\linewidth}
      \centering
      \includegraphics[width=\linewidth]{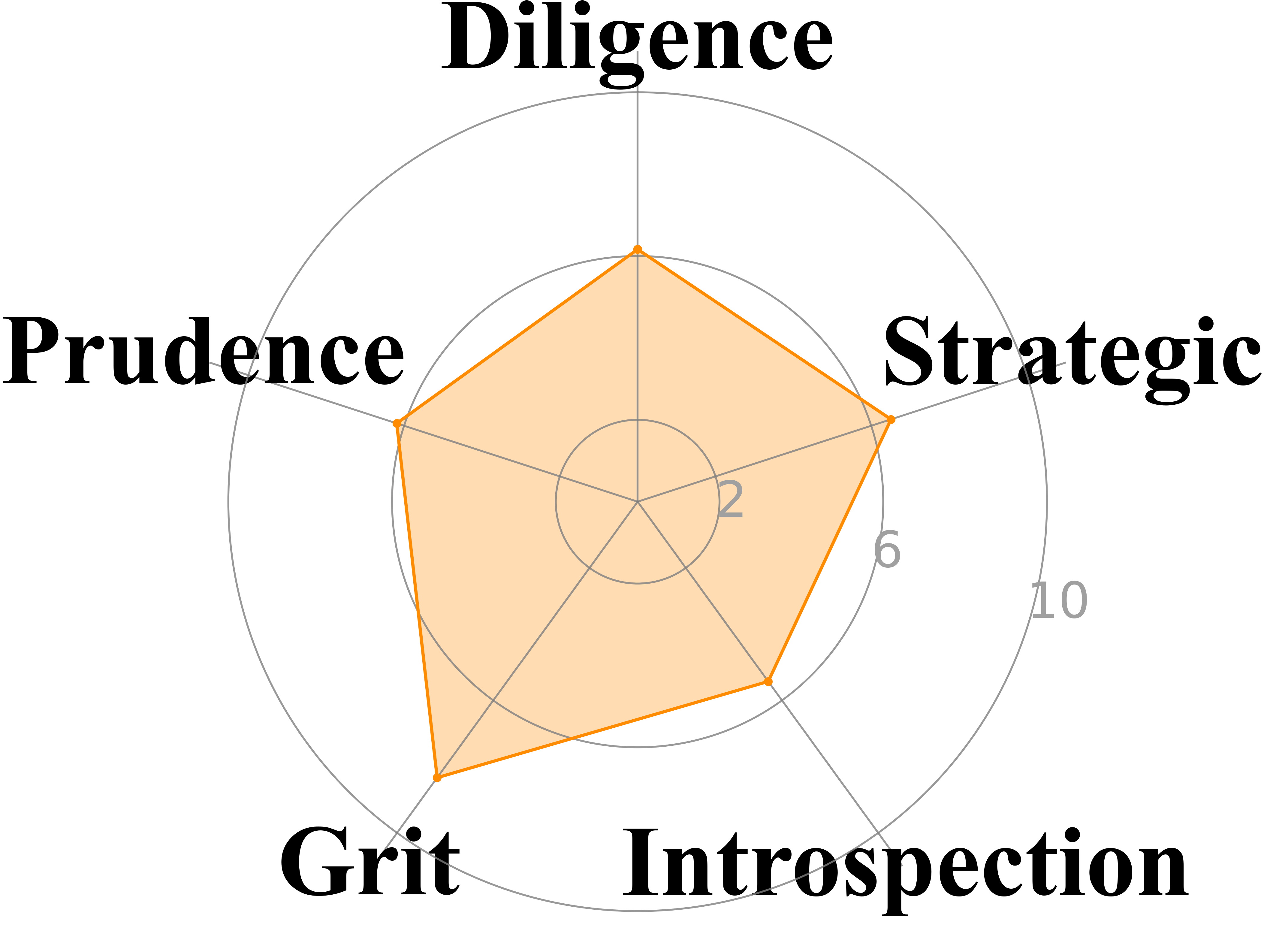}\vspace{1pt}
      \persona
        {DeepSeek-V3.2}
        {deepseekBG}
        {decisive executor}
    \end{minipage}

  \end{tabular}

  \caption{\small Five-dimensional personality radar charts of different MCP-based agents.}
  \label{fig:radar-3models}
\end{figure}
\subsection{Evaluation with Synthesized Environments}
\label{sec:eval-synthesized}

To validate the World Engine as a faithful substitute for live API execution, we run the same 50 evaluation tasks under the synthesized execution mode and compare the resulting model rankings against those obtained from real API execution. Table~\ref{tab:world-model-results} reports the per-model scores under synthesized execution powered by gemini-3.1-flash-lite-preview.

\begin{table}[t]
\centering
\small
\setlength{\tabcolsep}{4pt}
\caption{Model performance under synthesized execution via the World Engine. All tasks are executed without live API access.}
\label{tab:world-model-results}
\begin{tabular}{lcccc}
\toprule
Model & Overall & Pass\% & Compl. & Constr. \\
\midrule
claude-opus-4.5  & \textbf{0.869} & \textbf{66.7} & \textbf{8.6} & 9.0 \\
grok-4           & 0.858 & 56.8 & 8.7 & 8.7 \\
gpt-5.2          & 0.837 & 58.5 & 7.4 & 7.6 \\
gemini-3-pro     & 0.769 & 61.1 & 6.9 & \textbf{9.0} \\
deepseek-v3.2    & 0.764 & 38.9 & 7.0 & 8.6 \\
glm-4.6v         & 0.728 & 13.0 & 5.9 & 8.4 \\
qwen3-235b       & 0.714 & 14.8 & 5.8 & 8.4 \\
gpt-oss-120b     & 0.694 & 11.1 & 6.6 & 7.3 \\
gpt-4o-mini      & 0.483 & 0.0  & 2.9 & 4.1 \\
\bottomrule
\end{tabular}
\end{table}

To quantify ranking fidelity, we compute Spearman correlation between the model rankings produced by the World Engine and those from real API execution. The World Engine achieves $\rho = 0.883$ against the real-execution ranking, indicating that synthesized environments preserve the relative ordering of model capabilities with high fidelity. This confirms that the World Engine can reliably substitute live API calls for both large-scale trajectory generation and model evaluation, substantially reducing the cost and instability of real-service execution while maintaining diagnostic validity.

\subsection{Training with C-World}
\label{sec:training-results}

Beyond evaluation, C-World serves as a data engine for agent training. We curate trajectories from the 50 seed tasks defined in our experimental setup, isolating the most cognitively demanding phase: valid first-round actions where agents must translate abstract user intent into concrete tool search and invocation. This selection process yields a compact dataset of 1,170 samples that captures the core reasoning of tool selection rather than simple pattern matching.

Table~\ref{tab:main_results} compares models fine-tuned on C-World trajectories against baselines trained on significantly larger datasets. On the BFCL benchmark, models fine-tuned on C-World achieve top-tier accuracy (up to 30.05\%), surpassing baselines trained on datasets two orders of magnitude larger. On MCP-Universe, which evaluates generalization to real-world tools, C-World maintains robust performance across both architectures while baselines suffer degradation, confirming that the reasoning skills acquired from complex, constrained trajectories transfer effectively to unseen environments. Overall, C-World achieves best-in-class performance using only 1,170 samples, demonstrating extreme data efficiency enabled by the quality and realism of its generated trajectories.

\begin{table}[t]
\centering
\small
\setlength{\tabcolsep}{6pt}
\caption{Pass rates on BFCL and MCP-Universe. C-World achieves best-in-class performance with significantly less training data. \textit{Base} rows report zero-shot performance of the backbone before fine-tuning. Notably, zero-shot baselines on both backbones score below 20\% on BFCL and below 7\% on MCP-Universe, confirming that the gains reported here are attributable to fine-tuning rather than base capability.}
\label{tab:main_results}
\begin{tabular}{lccc}
\toprule
Model & Data Size & BFCL & MCP-Uni. \\
\midrule
\textit{Qwen2.5-7B} \\
\quad Base (zero-shot)   & --     & 19.93\% & 4.40\% \\
\quad + Toucan           & 119k   & 27.18\% & 15.28\% \\
\quad + ToolACE          & 11.3k  & 27.06\% & 2.23\% \\
\quad + \textbf{C-World} & \textbf{1.2k} & \textbf{28.58\%} & \textbf{15.30\%} \\
\midrule
\textit{Qwen3-8B} \\
\quad Base (zero-shot)   & --     & 18.32\% & 6.35\% \\
\quad + Toucan           & 119k   & 27.39\% & 6.67\% \\
\quad + ToolACE          & 11.3k  & 29.49\% & 3.29\% \\
\quad + \textbf{C-World} & \textbf{1.2k} & \textbf{30.05\%} & \textbf{8.86\%} \\
\bottomrule
\end{tabular}
\end{table}

\section{Related Work}
\paragraph{LLM-based Agents.}
Autonomous agent research has progressed from closed reasoning chains~\citep{wei2022chain} to ReAct-style reasoning--action loops~\citep{yao2022react}, from rigid function calling to Code-as-Action~\citep{wang2023voyager,wang2024codeact} and computer-using agents~\citep{zhouwebarena,xie2024osworld}, and toward self-correction~\citep{shinn2023reflexion,lin2023swiftsage} and multi-agent role specialization~\citep{li2023camel,hong2023metagpt} for long-horizon workflows.

\paragraph{Tool-Using Benchmarks.}
Early tool-use benchmarks focused on single-turn function calling~\citep{li2023apibank,patil2023gorilla} or multi-step chaining under pre-assigned toolsets~\citep{qin2023toolllm,mialon2023gaia,liu2023agentbench,patilberkeley,xu2024theagentcompany}. Recent efforts~\citep{fei2025mcpzero,mo2025livemcpbench} shift toward active tool retrieval from open registries, marking a transition from using to selecting and reasoning.

\paragraph{World Models for Agent Training.}
World models that learn environment dynamics for planning and data generation are well-studied in RL~\citep{ha2018world,hafner2023dreamerv3}. Recent work explores LLMs as zero-shot world models for text games~\citep{wang-etal-2024-language} and embodied settings~\citep{haijima2024embodied_icml}. We extend this to tool-use environments, simulating realistic tool behavior at the category level to enable API-free trajectory generation.

\section{Conclusion}
We presented C-World, an environment creation system defining agent environments through four components---Action Space, Task Distribution, Transition Function, and Reward Signal---with realistic and synthesized modes for both high-fidelity evaluation and scalable training. Evaluating nine state-of-the-art LLMs reveals that planning ability is uniformly strong while execution and constraint following remain the dominant bottlenecks, with DeepSeek-V3.2 emerging as the most robust open-weight contender. The World Engine achieves $\rho=0.883$ ranking correlation with real execution, and fine-tuning on just 1,170 C-World trajectories outperforms baselines trained on 119k samples, demonstrating C-World's dual value as a rigorous testbed and a data-efficient training engine.

Beyond its current instantiation, C-World's component-based design opens a path toward continuously evolving agent environments: as new MCP servers, task distributions, and failure modes emerge, C-World can integrate them without redesigning the underlying infrastructure. We hope C-World will serve as both a benchmark for understanding where today's agents fail and a scalable platform for training the next generation of tool-using agents.

\section*{Acknowledgments}
We sincerely thank all the anonymous reviewers for their insightful comments. We are deeply grateful to Henry Mao and the Smithery team for maintaining the MCP registry and for generously providing free access that made the curation and validation of 5,571 tools possible.

\section*{Limitations}
While C-World provides a scalable and realistic environment for training and testing tool-using agents under large tool pools, our current study has several limitations. Our evaluation set is limited: we synthesize 50 scenarios, which cannot exhaustively cover the combinatorial space of tool--server--constraint interactions. Although our round-robin sampling and fuzzy rewriting (§\ref{sec:task-distribution}) enforce cross-domain diversity within each scenario, and fine-tuning results on MCP-Universe (Table~\ref{tab:main_results}) empirically confirm out-of-distribution transfer, performance on rare domains or uncommon constraint combinations may still deviate from our leaderboard. Additionally, while we report small-model results in Appendix~\ref{app:small-models}, our main analysis focuses on frontier and large open models, and the behavioral patterns we identify (\eg the planning--execution gap) may manifest differently in the sub-10B regime.

\bibliography{custom}
\clearpage
\appendix

\section{Evaluation Metrics}
\label{app:eval-metrics}
To strictly evaluate open-world orchestration, we measure performance across four dimensions: 

\paragraph{Quality.}

(i) \emph{Completeness} quantifies the resolution of user intent to ensure the task is actually solved, measured by an LLM comparing the final response against the task requirements. 
(ii) \emph{Grounding} ensures factual reliability by an LLM verifier, checking that the answer is strictly derived from tool observation logs and penalizing any hallucinations not supported by the execution history.

\paragraph{Robustness.}
(i) \emph{Schema Compliance} ensures fundamental functionality by validating every tool call against the official JSON schema definitions to reject syntax or type errors.
(ii) \emph{Recovery Rate} measures self-correction by calculating the conditional probability of a successful execution in turn $t+1$ given a failure in turn $t$.
(iii) \emph{Flexibility} evaluates strategic adaptability by calculating the conditional probability that the agent takes an alternative execution plan after a State Controller–injected error, thereby identifying agents capable of adaptive replanning under failure.

\paragraph{Constraint.}
We evaluate adherence against the structured constraints pre-defined in the seed query:
(i) \emph{Order \& Diversity} ensure logical sequencing by comparing execution timestamps and server counts against the query's explicit dependency graph.
(ii) \emph{Format} guarantees structural compliance by using an LLM to verify that both the output structure and content match the patterns specified in the task requirements.
(iii) \emph{Tradeoff} evaluates reasoning using an LLM to verify if the resolution of conflicting objectives aligns with the implicit preferences embedded in the user request.

\paragraph{Planning.}
(i) \emph{Goal Decomposition} evaluates strategic alignment by employing an LLM judge to score the semantic overlap between the generated plan and the ground-truth reference rationales.
(ii) \emph{Progress Tracking} assesses state estimation accuracy by using an external evaluator to cross-check the Planner's status labels (\eg `Complete`) against the actual tool execution logs.
(iii) \emph{Efficiency} quantifies operational conciseness by recording the raw number of interaction turns required to successfully resolve the query.

We employ a unified scoring protocol: deterministic metrics (\eg Schema, Order) are reported as percentages(\%) over all evaluation instances, while semantic assessments (\eg Completeness, Tradeoff) are graded by the LLM Judge on a 10-point scale.

\section{Evaluated Models}
We evaluate 9 representative models (Table~\ref{tab:leaderboard_summary}) spanning proprietary and open-weight paradigms to form a holistic benchmark. This includes frontier closed-source models (\texttt{gpt-5.2}, \texttt{gemini-3-pro-preview}, \texttt{claude-opus-4.5}, \texttt{grok-4}), popular open-weight systems (\texttt{deepseek-v3.2}, \texttt{glm-4.6v}, \texttt{qwen3-235b-a22b}, \texttt{gpt-oss-120b}), as well as a cost-optimized baseline (\texttt{gpt-4o-mini}) to examine long-horizon reasoning abilities across mainstream models.

\section{Training Implementation}
We conduct experiments on Qwen2.5-7B-Instruct and Qwen3-8B to demonstrate the effectiveness and robustness of our dataset across different model architectures and scales.
To evaluate the advantages of our dataset, we compare it against Toucan~\citep{xu2025toucan} and ToolACE~\citep{liu2024toolace}.
All datasets are converted into a data format compatible with the ms-swift training framework.\footnote{\url{https://github.com/modelscope/ms-swift}}
Model training follows the Hermes-style agent supervision paradigm,\footnote{\url{https://huggingface.co/NousResearch/Hermes-2-Pro-Llama-3-8B}} which explicitly models tool usage through tool invocation and tool response messages. For baseline comparisons, we use the full ToolACE dataset containing 11.3K samples, and additionally incorporate a subset of Toucan data drawn from the SFT class, which comprises 119K trajectories in total.

\section{LLM Usage Statement}
Large language models (LLMs) were used solely for language editing—improving grammar, clarity, and overall readability. They did not generate or modify the manuscript’s scientific content, conceptual contributions, methodology, or experimental results. The authors assume full responsibility for the final text and its accuracy.

\section{Detailed Evaluation Results}
\label{app:detailed_results}

\begin{table*}[h!] 
\centering
\small
\setlength{\tabcolsep}{3pt}
\resizebox{\textwidth}{!}{%
\begin{tabular}{lcccccccccccccc}
\toprule
Model
& Overall
& \multicolumn{2}{c}{Answer Quality}
& \multicolumn{3}{c}{Tool Use Robustness}
& \multicolumn{4}{c}{Constraint Following}
& \multicolumn{4}{c}{Long-horizon} \\
\cmidrule(lr){3-4}\cmidrule(lr){5-7}\cmidrule(lr){8-11}\cmidrule(lr){12-15}
& Score & Completeness & Grounding & Success Rate & Recovery Rate & Flexibility &
Order & Info Diversity & Format & Tradeoff &
Tool Calls & \# Turns & Progress Tracking & Goal Decomposition \\

\midrule
\textbf{gemini-3-pro-preview} & 5.87 & 4.75 & 2.58 & 88.8\% & 89.0\% & 68.8\% & 53.8\% & 97.8\% & 53.9\% & 13.3\% & 47.86 & 3.20 & 7.60 & 8.66 \\
\textbf{claude-opus-4.5}      & 5.42 & 4.70 & 2.93 & 92.7\% & 83.7\% & 60.8\% & 65.4\% & 73.3\% & 51.0\% & 33.7\% & 45.16 & 4.01 & 6.41 & 7.72 \\
\textbf{deepseek-v3.2}        & 4.97 & 4.00 & 2.18 & 87.5\% & 90.6\% & 72.4\% & 73.1\% & 70.0\% & 39.5\% & 17.9\% & 21.73 & 4.92 & 6.46 & 8.04\\
\textbf{glm-4.6v}             & 4.86 & 4.01 & 1.18 & 84.8\% & 71.5\% & 57.3\% & 75.6\% & 52.2\% & 34.2\% & 11.5\% & 18.03 & 3.27 & 7.20 & 8.50 \\
\textbf{grok-4}               & 4.78 & 3.80 & 1.95 & 87.8\% & 89.0\% & 63.6\% & 64.1\% & 92.2\% & 68.3\% & 35.5\% & 27.37 & 2.55 & 6.02 & 8.28 \\
\textbf{gpt-oss-120b}         & 4.66 & 3.42 & 1.28 & 86.3\% & 72.7\% & 59.7\% & 87.2\% & 38.9\% & 35.8\% & 13.3\% & 14.40 & 3.14 & 6.53 & 8.10 \\
\textbf{gpt-5.2}              & 4.43 & 3.42 & 3.80 & 85.5\% & 79.3\% & 55.4\% & 71.6\% & 37.2\% & 12.4\% & 12.5\% & 29.20 & 2.30 & 5.62 & 7.73 \\
\textbf{qwen3-235b-a22b}      & 3.53 & 2.56 & 1.17 & 87.9\% & 88.1\% & 66.1\% & 80.8\% & 43.3\% & 31.3\% & 8.6\% & 11.15 & 4.41 & 6.93 & 8.51 \\
\textbf{gpt-4o-mini}          & 3.07 & 1.13 & 0.85 & 87.5\% & 50.6\% & 39.7\% & 85.9\% & 46.7\% & 3.3\% & 0.0\% & 51.71 & 6.45 & 6.00 & 7.71 \\
\bottomrule
\end{tabular}%
}
\caption{Full Leaderboard breakdown including sub-metrics for Answer Quality, Tool Use Robustness, Constraint Following, and Long-horizon planning.}
\label{tab:full_leaderboard_appendix}
\end{table*}

\section{State Controller Intervention Examples}
\label{app:state-controller-examples}

We illustrate each intervention category with representative examples drawn from evaluation logs.

\paragraph{Transient Infrastructure Anomaly (Triggering Retry).}
The agent calls \texttt{search\_legislation} on \texttt{@pma1999/boe} with \texttt{query="protección de datos personales"}, \texttt{solo\_vigentes=True}, \texttt{limit=10}, and receives \texttt{503 Service Unavailable}. It interprets the failure as a transient server-side issue and immediately retries the same tool with a narrower request by specifying \texttt{titulo="Reglamento (UE) 2016/679"} while keeping the other arguments. This illustrates that the agent continues on the same tool path under infrastructure instability, adjusting the request to be more targeted rather than switching domains.

\paragraph{Adversarial Tool Deprecation (Triggering Re-planning).}
The agent calls \texttt{search\_flights} on \texttt{@punitarani/fli} with \texttt{from\_airport="SFO"}, \texttt{to\_airport="JFK,LGA"}, \texttt{date="2026-02-10"}, and receives \texttt{404 Not Found}. It treats retrying as unproductive and switches to an alternative implementation, calling \texttt{search\_flights} on \texttt{@gvzq/flight-mcp} to continue the itinerary search. This demonstrates forced re-planning: the agent leaves an over-relied-upon tool and recovers via an alternative capability in the broader registry.

\paragraph{Payload Truncation (Triggering Refinement).}
The agent calls \texttt{get\_financial\_statements} on \texttt{@hollaugo/financial-research-mcp} with \texttt{ticker="LLY"} and receives a large JSON response ($\sim$36{,}535 chars). The controller truncates the payload to 30{,}000 chars, explicitly marking the truncation. The agent proceeds using the truncated observation as its available input, continuing the trajectory without access to the full payload.

\paragraph{Constraint-level Interruption (Triggering Re-optimization).}
After an initial response to an earnings-workflow request (AAPL, MSFT, NVDA, TSLA, AMZN, QQQ, SPY), the user introduces new constraints mid-episode: create a Google Sheet named ``Earnings Dashboard 2026,'' create January 2026 calendar events (``NVDA Deep Dive'' 90-min and ``AAPL Standard Review'' 45-min), create a Trello board ``Earnings Playbook'' with three specified lists, create a task ``Finalize QQQ ETF Memo'' due 3 days after the QQQ earnings date, and rename the Contentful space to ``Earnings Research Hub.'' These constraints partially invalidate the prior plan without resetting the episode. The agent must preserve valid progress while re-optimizing the remaining steps to satisfy the updated requirements.

\paragraph{Fairness Guarantee.}
To ensure the adversarial injection does not create unequal testing conditions, the exact turn at which a given intervention fires is randomized across models, but the total frequency of each intervention type is held strictly constant across all evaluated agents. Every model faces the same ``adversity budget,'' so our leaderboard reflects genuine recovery capability rather than uneven stress.

\section{Per-Event-Type Robustness Breakdown}
\label{app:event-breakdown}

Table~\ref{tab:event-breakdown} decomposes aggregate Recovery Rate and Flexibility into the three main intervention categories introduced by the state controller. The breakdown exposes behavioral distinctions that aggregate metrics obscure: for timeouts, where retrying is the correct response, top models such as DeepSeek-V3.2 achieve perfect recovery; for unavailability, where retrying is futile and the agent must switch tools, stronger planners maintain high flexibility while weaker models collapse. State drift---where the observation is delivered but semantically altered---is uniformly the hardest to recover from, with flexibility dropping across all models.

\begin{table}[t]
\centering
\small
\setlength{\tabcolsep}{3pt}
\caption{Per-event-type recovery rate / flexibility (\%). Each cell shows Recovery Rate~/~Flexibility under the specified intervention category.}
\label{tab:event-breakdown}
\begin{tabular}{lccc}
\toprule
Model & Timeout & Unavailability & State Drift \\
\midrule
DeepSeek-V3.2   & \textbf{100.0} / \textbf{84.9} & \textbf{94.7} / \textbf{84.3} & 93.5 / \textbf{54.9} \\
Claude-Opus-4.5 & 88.3 / 45.9  & 89.3 / 76.9 & 84.9 / 50.0 \\
Gemini-3-Pro    & 83.3 / 71.4  & 82.9 / 84.4 & \textbf{98.3} / 51.0 \\
GPT-5.2         & 61.9 / 69.6  & 69.2 / 71.1 & 98.1 / 29.5 \\
GPT-4o-mini     & 41.7 / 52.9  & 55.9 / 36.6 & 55.1 / 33.1 \\
\bottomrule
\end{tabular}
\end{table}

\section{Evaluation on Small Models}
\label{app:small-models}

To complement our frontier-model evaluation, we additionally test representative models below 10B parameters on the same C-World benchmark. Table~\ref{tab:small-models} reports overall scores. As expected, compact models trail frontier systems substantially, but relative ordering among them is informative: QwQ-32B's reasoning-oriented training yields the strongest small-model performance (4.55), outperforming several larger models in our main table. Among sub-10B candidates, Qwen-2.5-7B-Instruct (3.00) leads, followed by gemma-3-4b-it (2.54) and Llama-3-8B-Instruct (2.12). These results confirm that the C-World benchmark remains discriminative in the compact regime and that general instruction tuning alone is insufficient for long-horizon tool use at this scale.

\begin{table}[t]
\centering
\small
\setlength{\tabcolsep}{10pt}
\caption{Overall C-World scores for compact models ($\leq$32B parameters).}
\label{tab:small-models}
\begin{tabular}{lc}
\toprule
Model & Overall \\
\midrule
QwQ-32B                & \textbf{4.55} \\
Qwen-2.5-7B-Instruct   & 3.00 \\
gemma-3-4b-it          & 2.54 \\
Llama-3-8B-Instruct    & 2.12 \\
\bottomrule
\end{tabular}
\end{table}

\section{Human Alignment Measurement}
\label{app:human-alignment}

We assess the stability of our LLM evaluation on $N=40$ queries (Table~\ref{tab:agreement_combined}). Concretely, we ask human annotators to independently rank anonymized trajectories, and we compute Spearman correlations for both human--human and human--judge agreement. The resulting average human--human agreement is $\rho = 0.773$. In comparison, DeepSeek-V3.2 ($\rho = 0.759$) and GPT-5.1 ($\rho = 0.733$) exhibit close alignment with the human consensus. Although GPT-4o attains a lower mean correlation ($\rho = 0.688$), the uncertainty ranges of human--human and human--LLM correlations substantially overlap when accounting for their respective standard deviations. Together, these results demonstrate the human-level reliability and robustness of the C-World scoring framework.

\begin{table}[h]
\centering
\small
\setlength{\tabcolsep}{6pt}
\begin{tabular}{l c l c}
\toprule
Left & vs. & Right & $\rho$ (avg. $\pm$ std.) \\
\midrule
Human          & vs. & Human  & $0.773 \pm 0.075$ \\
DeepSeek-V3.2  & vs. & Human  & $0.759 \pm 0.073$ \\
GPT-4o         & vs. & Human  & $0.688 \pm 0.109$ \\
GPT-5.1        & vs. & Human  & $0.733 \pm 0.091$ \\
\bottomrule
\end{tabular}
\caption{Comparison of human--human agreement and human--LLM alignment measured by Spearman $\rho$ on the same set of 40 queries.}
\label{tab:agreement_combined}
\end{table}

\paragraph{Robustness to Judge Family Bias.}
A natural concern is that including two OpenAI judges (GPT-4o, GPT-5.1) in our three-judge ensemble might bias scores toward OpenAI models. The leaderboard in Table~\ref{tab:leaderboard_summary} directly refutes this: GPT-4o-mini ranks last overall, and GPT-5.2 is outperformed by Gemini-3-Pro, Claude-Opus-4.5, and DeepSeek-V3.2. The cross-family ensemble combined with majority-vote aggregation thus successfully neutralizes single-developer preferences. Moreover, the observed human--LLM Spearman correlations ($\rho \approx 0.73$--$0.76$) closely approach the human--human ceiling of $\rho = 0.773$, suggesting that remaining disagreement reflects the inherent subjectivity of evaluating 28.5-turn trajectories rather than judge unreliability.

\section{Characteristic Analysis: Dimension Definitions}
\label{app:personas}

The five anthropomorphic dimensions used in Figure~\ref{fig:radar-3models} are defined as follows:

\begin{itemize}
    \item \emph{Diligence}: commitment to executing every subgoal, measured by the extent of reasoning steps per turn.
    \item \emph{Prudence}: strict reliance on external verification, measured by the rarity of triggering the ``no-tool-for-3-turns'' termination condition.
    \item \emph{Grit}: resilience in adversity, defined as the combination of robustness and flexibility under injected failures.
    \item \emph{Introspection}: persistence in extending dialogue depth to chase down missing details rather than giving up early.
    \item \emph{Strategic}: preference for explicit task decomposition, measured by the proportion of plans with multi-step sub-goal graphs.
\end{itemize}

During our open-world evaluations, we observe distinct decision-making styles across models. For instance, gemini-3-pro-preview behaves like a \emph{meticulous scholar}, choosing to expend extensive acting-reasoning steps to rigorously verify every subgoal before moving forward, whereas gpt-4o-mini is more \emph{impulsive}, often rushing into tool loops without grounding its actions and frequently triggering stagnation checks. In contrast, deepseek-v3.2 operates as a \emph{decisive executor}, maintaining high quality with minimal overhead across trajectories.
\end{document}